\documentclass[sigconf]{acmart}
\renewcommand\footnotetextcopyrightpermission[1]{} % removes footnote with conference information in first column
\makeatletter
\renewcommand\@formatdoi[1]{\ignorespaces}
\makeatother

\AtBeginDocument{%
  \providecommand\BibTeX{{%
    \normalfont B\kern-0.5em{\scshape i\kern-0.25em b}\kern-0.8em\TeX}}}

\usepackage{subfigure}
\usepackage{float} 
\usepackage{hyperref} 
\usepackage{url} 
\usepackage{verbatim} 
\usepackage{color} 
\usepackage{multirow} 
\usepackage{color} 

\begin{document} 
\sloppy 

%%
%% The "title" command has an optional parameter,
%% allowing the author to define a "short title" to be used in page headers.
\title{Action-based Early Autism Diagnosis Using \\
Contrastive Feature Learning} 
%with Scarce Data} 

%%
%% The "author" command and its associated commands are used to define
%% the authors and their affiliations.
%% Of note is the shared affiliation of the first two authors, and the
%% "authornote" and "authornotemark" commands
%% used to denote shared contribution to the research.
\author{Asha Rani}
%\orcid{1234-5678-9012}
\affiliation{%
  \institution{IIT Jodhpur, India}
  %\country{India}
}
\email{rani.1@iitj.ac.in}

\author{Pankaj Yadav}
\affiliation{%
  \institution{IIT Jodhpur, India}
  %\country{India}
  }
\email{pyadav@iitj.ac.in}

\author{Yashaswi Verma}
\affiliation{%
  \institution{IIT Jodhpur, India}
  %\country{India}
}
\email{yashaswi@iitj.ac.in}

%%
%% By default, the full list of authors will be used in the page
%% headers. Often, this list is too long, and will overlap
%% other information printed in the page headers. This command allows
%% the author to define a more concise list
%% of authors' names for this purpose.
\renewcommand{\shortauthors}{Rani, et al.}
%\renewcommand{}{}
%%
%% The abstract is a short summary of the work to be presented in the
%% article.
\begin{abstract} 

Autism, also known as Autism Spectrum Disorder (or 
ASD), 
is a neurological disorder. Its 
main symptoms include difficulty in 
(verbal and/or non-verbal) communication, 
and rigid/repetitive behavior. 
These symptoms are often indistinguishable from a normal (control) individual, due to which this 
disorder remains undiagnosed in early childhood, 
leading to delayed treatment. 
Since the learning curve is steep during the initial years, an early diagnosis of autism would allow to make an early intervention, which would positively affect the growth of an autistic child. 
Further, the traditional methods of autism diagnosis require multiple visits to a specialized doctor, however this process is generally time-consuming. 
In this paper, we present a learning based approach 
to automate autism diagnosis using simple and 
small action video clips of subjects. 
This task is particularly challenging because 
the amount of annotated data available is small, and 
the variations among samples from the 
two categories 
(ASD and control) are generally indistinguishable. 
This is also evident from poor performance of a binary classifier learned using the cross-entropy loss on top of a baseline deep encoder. 
To address this, we adopt contrastive feature learning in both self-supervised and supervised 
learning frameworks, and show that these lead to a 
significant increase in the 
prediction accuracy of a binary 
classifier on this task. 
We further validate this 
by conducting thorough experimental analyses under 
different set-ups on two publicly available 
datasets. 
Our code and pre-trained models 
are available \href{https://github.com/asharani97/CLRE_autism}{\textcolor{black}{here}} for reproducibility. %and wider usage. 
\footnote{“This preprint has not undergone peer review (when applicable) or any post-submission improvements or corrections. The Version of Record of this article is published in Multimedia Systems, and is available online at \href{https://doi.org/10.1007/s00530-023-01132-8}{\textcolor{black}{https://doi.org/10.1007/s00530-023-01132-8}}”.}
\end{abstract}

%%
%% The code below is generated by the tool at http://dl.acm.org/ccs.cfm.
%% Please copy and paste the code instead of the example below.
%%


\maketitle
%\fancyfoot{}
\pagestyle{plain}
%------------------------ Introduction ----------------------------------------------

\section{Introduction}
Contrastive feature learning has been used 
in various machine learning tasks in the recent 
times~\cite{Chen2020SimCLR,NEURIPS2020SupCon}. 
The main advantage of this approach is its ability to learn in a discriminative manner, %generate more data from the available dataset, 
which is beneficial when the availability of 
labelled data is limited. This is particularly 
the case in medical domain where the 
the process of data annotation is quite costly and 
time consuming. 
Autism is one such medical 
condition where it is challenging to collect and 
reliably annotate data, thus leading to availability of small datasets. 
Autism, also known as Autism Spectrum Disorder 
(or ASD), is a neurological 
disorder which largely affects an 
individual's cognitive capabilities, with 
the main symptoms being difficulty in social 
interaction and rigid behavior. 
The number of individuals suffering from 
autism has increased over the last few year 
\cite{stats}. 
Due to this, its diagnosis 
at an early stage can be 
helpful in early intervention. 
Traditionally, the diagnosis has been done by 
trained medical practitioners, which can be time-consuming and may delay an early diagnosis and intervention. 

In this work, our objective is to perform autism 
diagnosis in an automated manner. 
There are some recent machine learning based methods 
which were proposed to automate the ASD diagnosis process. 
These works have investigated data from different 
modalities, such as 
%Most of the work is concentrated towards dealing with single modality based solutions. The different modalities/ aspects which are investigated include 
eye tracking data, gesture videos, MRI data, 
EEG data, etc. \cite{Liu2016face,Jiang2017saliency,Ruan2021phototaking,ICPR2018Gesture,Tian2019VideoBasedEA,Sun2020SAB3D,Tawhid2021EEG,ref7,Heinsfeld2018IdentificationOA,Sherkatghanad2019AutomatedDO,Kong2019ClassificationOA}. 
Earlier analysis revealed that individuals with autism have an atypical sight \cite{Dawson2005atypical}. 
For example, \cite{WANG20153layeredmodel,Jiang2017saliency} worked on this objective, 
with~\cite{Jiang2017saliency} focusing on 
visual bias towards different objects, contrast and colour, and~\cite{WANG20153layeredmodel} focusing on 
using these aspects to predict ASD and control 
subjects using deep visual models like VGG16 \cite{2015vgg}. 
A different perspective to distinguish the two 
categories was introduced in \cite{Ruan2021phototaking} where the  first-person view of a scene (or visual) 
of 
an individual is compared for analysis.
A few works have used eye tracking data to distinguish on the basis of attention \cite{Tian2019VideoBasedEA}, 
hand gesture data (small action clips)  \cite{ICPR2018Gesture,Sun2020SAB3D}, 
EEG signals \cite{Tawhid2021EEG,BAYGIN2021104548,9344596} and MRI data \cite{Heinsfeld2018IdentificationOA,Kong2019ClassificationOA}. 

%used hand gesture data (small action clips) LSTM model, further an improvement of the results of this paper done by using spatial information on the same dataset \cite{Sun2020SAB3D}. 

%\par
%In \cite{Tawhid2021EEG,BAYGIN2021104548,9344596} authors used EEG signals, converted to workable form to use this as a aspect for distinguishing. In \cite{ref7,Heinsfeld2018IdentificationOA,Sherkatghanad2019AutomatedDO,Kong2019ClassificationOA} authors investigate MRI image based information to study on ASD, all focusing on some different machine learning methods. 

\begin{figure}[h]
\centering 
\begin{tabular}{c} 
\includegraphics[width=\linewidth,scale=0.1]{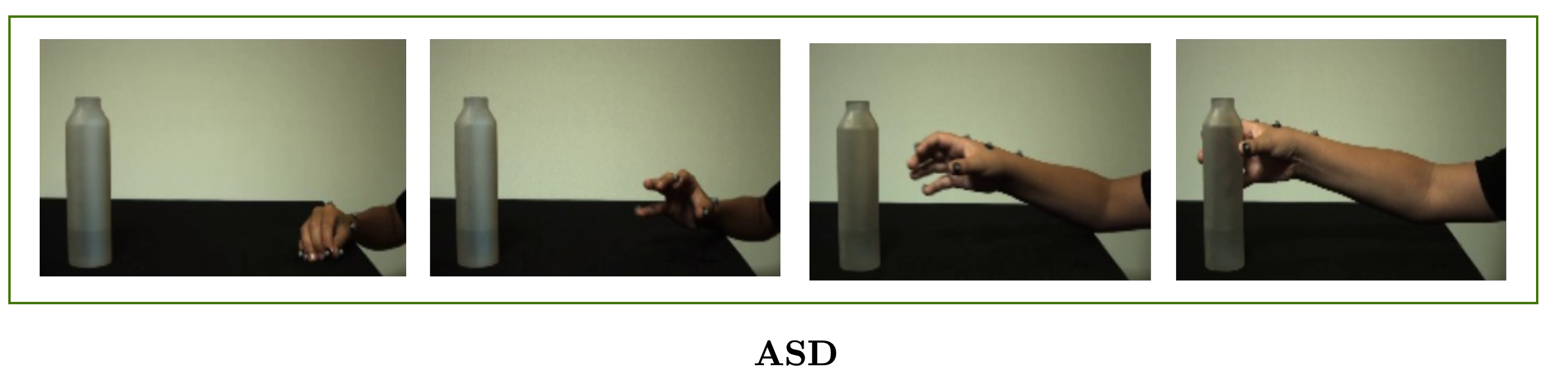}
\\ 
\includegraphics[width=\linewidth,scale=0.1]{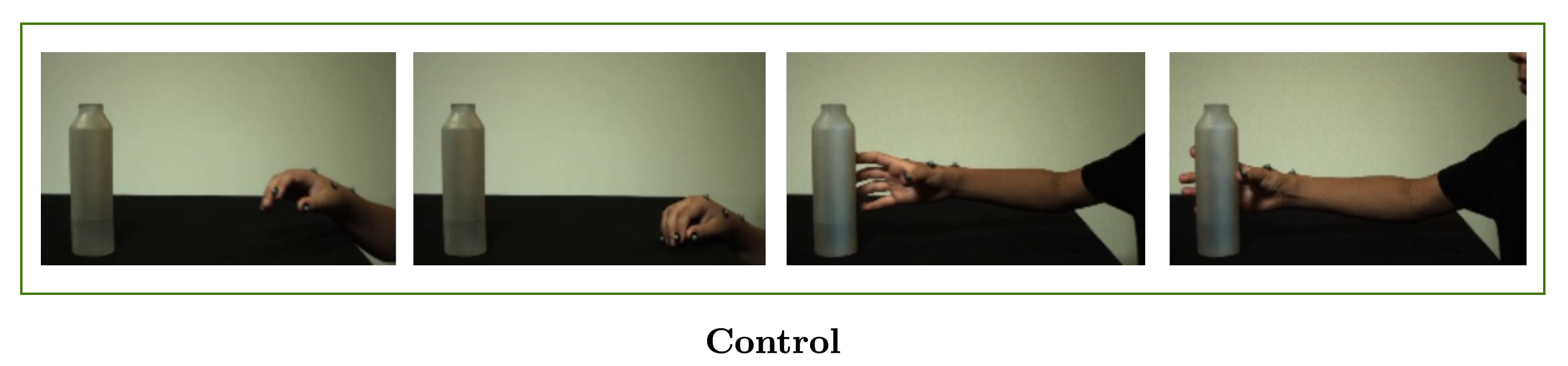}
\end{tabular}
\caption{Sample frames from videos of the two 
categories (ASD and Control) from the Hand 
Gesture Dataset (HGD)~\cite{ICPR2018Gesture}. 
Note that the two categories are visually 
indistinguishable, thus leading to difficulty 
in learning discriminative features.} 
\label{fig:fig1}
\end{figure}

\begin{comment} 
\begin{figure}[h]
\centering 
\begin{tabular}{c} 
\begin{subfigure}[]
    {\includegraphics[width=\linewidth,scale = 0.1]{Sample_Images/asd_sample.png}
}
\end{subfigure}
\hfill
\begin{subfigure}[]
    {\includegraphics[width=\linewidth,scale=0.1]{Sample_Images/control_sample.png}
}
\end{subfigure} 
\end{tabular}
\caption{Sample frames from videos of Hand Gesture Dataset (HGD)} \cite{ICPR2018Gesture} 
\end{figure} 
\end{comment} 
  
In this paper, we approach automated autism 
diagnosis using action-based color/gray-scale 
videos. 
This is because such data is much easier to acquire 
compared to other modalities such as EEG, MRI or 
eye-tracking data, and can also be readily 
used for training deep learning based 
models. 
However, it is difficult to obtain good prediction 
accuracy using such data because it 
exhibits small inter-class variability if 
acquired in a flexible/less-controlled 
environment. 
E.g., Figure~\ref{fig:fig1} shows a sample from both 
ASD and control categories of 
one such dataset (the Hand Gesture Dataset\cite{ICPR2018Gesture}). 
We can observe that both the samples look quite identical, and if we attempt to differentiate them just by seeing them, it would be nearly impossible to identify the correct category of a sample. 
Because of this, 
the conventional deep learning models used in visual 
classification tasks may not be directly adopted for 
this task. This is also evident from our experimental 
analyses where a binary classifier trained on top 
of a deep network gives poor performance. 
To address this, we propose to use 
constrastive feature learning to learn 
minute differences between the two categories. 
It is also important to note that our application of 
contrastive feature learning is quite different from 
the conventional Computer Vision tasks since 
(a) our datasets contain a small number of samples, 
(b) the data is inherently complex being video data, 
and (c) the samples within a dataset exhibit 
low inter-class variability and are nearly 
impossible to distinguish visually 
by a normal human being. 
We show experimentally that contrastive feature 
learning can be quite beneficial in such tasks, and 
the features thus learned lead to a significant 
increase in the prediction accuracy of a binary 
classifier. 

In the next section, we discuss the two 
contrastive feature learning techniques we have adopted in this work. In section~\ref{sec:experiment}, 
we provide the experimental details and 
discuss our empirical findings under 
different experimental set-ups, and then we conclude in section~\ref{sec:Conclusion}.

%We discuss the methods used in the paper in section  \ref{sec:methods}, various experiments  where we show how binary classification is insufficient for such a task where variation between the two categories is minimal, and data samples are also pretty low at the same time in \ref{sec:experiment} and also we show our experimentation with different data setups using self-supervised contrastive learning and compare them with the binary classification in \ref{sec:results} and conclusion in the end.
 
%--------------------------Method----------------------------------

\section{Contrastive Feature Learning}
\label{sec:methods}
Below we discuss the two contrastive 
feature learning techniques we have adopted in this 
work. Specifically, we perform 
contrastive learning in both 
self-supervised~\cite{Chen2020SimCLR} and 
supervised~\cite{NEURIPS2020SupCon} set-ups. 
Lately, these techniques have been found to be 
useful in learning discriminative features by 
making use of additional data 
generated through 
minor variations in the original data, and then creating pairs of similar and 
dissimilar samples 
to learn discriminative features. 
%have used contrastive learning methods SimCLR \ref{subsec:SimCLR} and SupCLR \ref{subsec:SupCLR} as it generates more data during the training process in order to learn the representation better, and can also work well on such complex data. 
For consistency, in the following sub-sections, 
we denote the base 
network (encoder) using the function $e()$, 
and a non-linear 
multi-layer perceptron (MLP) layer using the 
function $h()$. 

%We are using following notation in the upcoming section, encoder/base network is denoted using $e()$. While non-linear MLP layer used further is denoted as h().

\subsection{Self-Supervised Contrastive Learning}
\label{subsec:SimCLR}
Contrastive learning generally requires a memory bank or a specialized architecture to learn the visual representations of objects. This leads to an increased memory requirement and/or a complex training procedure. To address this, in \cite{Chen2020SimCLR}, the authors proposed a self-supervised contrastive learning 
technique (SimCLR) which does not require any specialized architecture or memory bank. Instead, 
%The paper's primary idea is in the way 
it integrated various existing and simple 
approaches in the literature for self-supervised learning, and achieved state-of-the-art results on the 
image classification task. 
Specifically, this approach uses simple 
transformations (e.g., vertical flip, addition of a 
small noise, etc.) to generate new samples, and uses 
them to create positive and negative pairs. 
A positive pair is created using a given 
sample and its transformed version, and a negative 
pair is created by combining a given sample 
with another sample. % in a batch. 
%Using these pairs, the encoder is trained using a normalized contrastive loss thus learning discriminative features. 
%The self-supervised contrastive learning uses unlabelled data and transformation is used for increasing the dataset size and generating positive pairs. It uses both the transformed as well as the original data to learn the similarity. 
%This minor change helps in better representation learning. The samples taken from the same class (\ie, a given sample and its transformed version) are considered as positive pairs, while the pairs containing samples from different classes (\ie, two distinct samples) are considered as negative pairs. 
During training, both the samples in a pair are passed through the same encoder $e()$, and the features obtained are then passed to a non-linear projection head $h()$ which helps in improving the discriminative ability of the learned features. %encoder $e()$. 
The whole network is trained using the 
following loss, 
%The loss function that is computed is called the $NT-Xent$ loss, 
which results in updating both 
$e()$ and $h()$ simultaneously: 
%. The training process updates the functions $e()$ and $h()$, 
\begin{equation}
  \ell_{i,j} = -\log\frac{\exp(\text{sim}(z_{i},z_{j})/\tau)}{\sum _{k=1,k\neq i}^{2N}%\mathbbm{1}_{k\neq i}
  \exp(\text{sim}( z_{i},z_{k})/\tau)}
  \label{eq:lossfun}
\end{equation}
Here, the subscripts $i,j,k$ refer to an anchor point, a positive sample and a negative sample, $\text{sim}()$ denotes cosine similarity between a pair of samples, $z$ denotes the output feature obtained from $h()$, $\ell_{i,j}$ denotes the loss for the 
given positive pair, and 
%from the last step and 
$N$ denotes the number of samples in a mini-batch. The whole network is trained using a gradient descent 
method. 
During the testing phase, %from which 
$h()$ is dropped and only 
$e()$ is used for feature 
extraction. 
%classification and $h()$ is not used during the testing phase. 
On top of $e()$, a classification layer is learned using the 
available labelled data. 
%At the end, the model performs classification using a linear module on top of 
%$e()$. 

An important advantage of this technique is that it allows learning discriminative 
representations of data 
points which are closely related without needing labelled data. 
Another important aspect, unlike other contrastive learning techniques, is that 
its training step 
uses a global batch normalization ({\it i.e.}, the 
mean values used for batch normalization are calculated locally), 
%to use information leakage), 
which subsequently helps 
%. The information leakage helps for 
in training the network 
%achieving good classification accuracy 
without requiring a memory bank. 
%without the need for better representations.
%The loss function of SimCLR \cite{Chen2020SimCLR} is as follows:

\subsection{Supervised Contrastive Learning}
\label{subsec:SupCLR}
In \cite{NEURIPS2020SupCon}, the authors extended the above self-supervised contrastive learning (SupCLR) technique 
%of \cite{Chen2020SimCLR} 
by introducing supervision in the form of 
labels available in the training data. 
%by introducing labels 
%to self-supervision setup i.e  as Supervised Contrastive Learning. 
This results in supervised contrastive learning, 
where the label information is used to identify 
{\it multiple} positive samples for a given sample in a batch. 
%The most of the parts remain same in the architecture, only in supervised setting the labels are also passed along with the samples, which is further used in the loss function to identify all the positive samples in the batch. 
%The modified loss used in this paper is named as $SupCon$ loss. 
Unlike the self-supervised learning set-up where the loss is calculated based on a given sample and its transformed 
version, it uses {\it all} the %positive pairs i.e the original and the augmented version. Here 
positive samples to create positive pairs ({\it i.e.}, the samples that belong to the same category) in a mini-batch. 
%comparison is done with all of the positive samples in the mini-batch. 
Because of this, while there is only one positive 
pair in case of self-supervised contrastive learning, 
there may be multiple such pairs in case of supervised contrastive learning, whose  
%setting only one positive pair is used while in supervised setting it includes all the positive pair, and hence improving the loss function further. 
loss for a given 
sample %in a mini-batch is  
is computed as follows: 
\begin{equation}
l_{i}^{sup} = \frac{-1}{|P( i) |} \sum_{p\ \in P(i)} \log\frac{\exp(z_{i}\cdot z_{p}/\tau)}{\sum _{a \in A} \exp(z_{i} \cdot z_{a}/ \tau)}
\label{eq:supconlossfun}
\end{equation} 
%\begin{equation}
%\sum _{i\ \in I} L_{out,i}^{sup} = \sum _{i\ \in I}\frac{-1}{|P( i) |} \sum_{p\ \in P(i)} \log\frac{\exp(z_{i}\cdot z_{p}/\tau)}{\sum _{a \in A} \exp(z_{i} \cdot z_{a}/ \tau)}
%\label{eq:supconlossfun}
%\end{equation} 
where $P(i)$ denotes the set of indices of 
positive samples 
with respect to the $i^{th}$ sample in a mini-batch, and $A$ denotes the set of indices 
in the mini-batch excluding $i$. 
%$z_{p}$ indicates all the positive samples in the mini-batch i.e samples belonging to the same label class. Here $i \in I = \{1,2,3 .. N\}$. The set $A(i) \equiv I \setminus \{i\} $ , while $, P(i) \equiv {p \in A(i) : y_{p} = y{i}} $ where $y$ indicates label.

In~\cite{NEURIPS2020SupCon}, it was shown that a deep network trained with the supervised contrastive loss achieves better 
accuracy than what it achieves when trained 
using the self-supervised loss. 
%In paper, results show that providing contrastive learning samples with label information works well. The authors claimed to perform better with the same architecture used in SimCLR. 
The authors suggest that this is because in the process of representation learning, clustering of data points improves in the presence of labels, as it avoids data points from different categories to come closer. 
%The modified loss function had the ability to perform hard positive or negative mining. 

\section{Experiments}
\label{sec:experiment}
\subsection{Datasets}
%Our dataset selection as mentioned earlier consist of samples where variation between the two categories is very less. 
%\newline
We have used two datasets to evaluate and analyse the above methods for the Autism diagnosis task: Hand Gesture Dataset \cite{ICPR2018Gesture} and Autism Dataset \cite{AAAI2020GuidedSupervision}. In both the datasets, subjects are asked to perform some predefined action in a controlled environment. 

\noindent 
\textbf{Hand Gesture Dataset (HGD)}: Hand Gesture Dataset is a video dataset that contains actions from four classes: Placing, Pouring, Pass to Place and Pass to Pour. The video-clips are recorded by asking 39 individuals (19 ASD and 20 Control) to perform each of the above four actions. Figure~\ref{fig:fig1} shows two samples from the HGD dataset. As discussed earlier, this is a very challenging dataset with limited samples and extremely low inter-class %(Autism versus Control) 
variability. 

%\begin{figure}[!h]
%\centering
%\includegraphics[width=\linewidth,scale = 0.1]{Sample_Images/hand_Sample2.png}
%\caption{Sample frames from videos of Hand Gesture Dataset (HGD)} \cite{ICPR2018Gesture}
%\label{fig:fig1}
%\end{figure}

\noindent 
\textbf{Autism Dataset (AD)}: This is also a video dataset, with videos from eight action classes: move the table, touch ear, lock hands, touch head, touch nose, rolly polly, tapping, and arms up. Unlike the HGD dataset, it has samples only from the ASD class. The above eight actions are present in two types of clips: tutor facing and child facing . 
To get samples from the control class, we have added videos from the publically available  HMDB51 dataset \cite{Kuehne11} following an earlier work \cite{AAAI2020GuidedSupervision}.
After merging the HMDB51 dataset, we call the new dataset obtained as ``AD+CD'' dataset. Note that unlike the HGD dataset, the 
AD+CD dataset is imbalanced as the number of ASD samples are more compared to the Control samples. 
%\textcolor{red}{
Further, since the samples in the two categories of the AD+CD dataset come from separate datasets with different background settings, their distributions are significantly different. This will also be evident in our empirical results where all the compared methods achieve near-perfect accuracy on this dataset. 
%} 
%Figure~\ref{fig:fig2}, shows two samples from the Autism Dataset. 
%\footnote{Both the datasets used in our experiments were shared by their original authors upon request.} 
%\begin{figure}[!h]
%\centering
%\includegraphics[width=\linewidth,scale = 0.1]{Sample_Images/autsim_dataset.png}
%\caption{Sample frames from videos of Autism Dataset, Rolly Polly (left) and Arms Up (right) %\cite{AAAI2020GuidedSupervision}.}
%\label{fig:fig2}
%\end{figure}

%-----------------------------------------------------------%

\subsection{Network Architecture} 

As we are working with video data, 
for both self-supervised and supervised contrastive learning approaches, we use 
R(2+1)D \cite{2018_r2plus1d} 
as the base network (or encoder $e()$).  
%The $e()$ used is an 18-layer R(2+1)D network. 
A set containing a fixed number of frames ({\it i.e.,} a clip of a fixed length) 
%(frames of a specified clip length) 
is passed through the encoder $e()$, along with an augmented version of the frames. 
The output of the encoder is an intermediate 512-dimensional representation. % $z$. 
This is then passed to a non-linear MLP layer $h()$ which gives a 
256-dimensional representation as the output. 
For classification purpose, we drop the MLP layer $h()$ and use only the learned encoder $e()$. 
%and add a linear classification layer after it. 
%For evaluation, a linear layer (classifier) is appended to the learned $e()$. 
To compare the two contrastive feature learning approaches with a baseline binary classifier (BinClassifier), we use the trained baseline encoder network $e()$ and add a linear classifier on top of it, and then re-train the entire network in an end-to-end manner.

%-----------------------------------------------------------%

\subsection{Training Details} % \& Evaluation}
\label{subsec:training}
%In the presented experiments, 
As part of data pre-processing, 
%we first extract frames in a sequential manner. 
each video is loaded in the form of a sequence of frames, and then we pick 16 frames in case of the HGD dataset and 10 frames in case of the AD+CD dataset following uniform sampling. To create pairs in self-supervised contrastive learning, %for the purpose of creating pairs, 
horizontal flip is used for generating a transformed (positive) sample. We use the following values of hyper-parameters during its training: 
%for SimCLR : 
learning rate ($lr$) of  $1e^{-4}$, weight decay ($w_{d}$) of $1e^{-6}$, and the Adam optimizer. 
In supervised contrastive learning, these values are: $lr=1e^{-3}$, $w_{d}=1e^{-6}$, momentum ($m$)$= 0.9$ and the SGD optimizer. 
For the baseline binary classifier, %In binary classification, 
the binary cross-entropy loss is used with the Adam optimizer, and $lr$ and $w_{d}$ are the same as in supervised contrastive learning. 
%, and the optimizer used is Adam. 
%In the first stage feature extraction is done using contrastive loss for SimCLR and SupCLR. This can be said as representation learning, where similar are attracted while different / dissimilar features are repelled. Then by using the saved model evaluation is carried out. The code used for the implementation: \url{https://drive.google.com/drive/folders/1eElkMQtTpvtXFjeMHbjNzZEU_xcM5S3P?usp=sharing}

%%%--extra---

%--------------------------------------------------------% 

%%%%-----------Classwise Table---------------------
\begin{table}[t] %[!h]
  \centering 
  \begin{tabular}{@{}lc@{}lc@{}rc@{}}
    \toprule
    \multicolumn{1}{c}{{\bf Hand Gesture Dataset}} & \multicolumn{3}{c}{Accuracy \%}  \\
    \midrule
    \multicolumn{1}{c}{Action Class} & \multicolumn{1}{c}{BinClassifier}& \multicolumn{1}{c}{SimCLR} & \multicolumn{1}{c}{SupCLR}\\
    \midrule
    \multicolumn{1}{c}{Pass to Place} & \multicolumn{1}{c}{51.09} &
    \multicolumn{1}{c}{{70.07}} & \multicolumn{1}{c}{\textbf{70.80}
    }\\
    \multicolumn{1}{c}{Pass to Pour}  & \multicolumn{1}{c}{54.23}& 
    \multicolumn{1}{c}{\textbf{73.24}} & \multicolumn{1}{c}{69.72}  \\
    \multicolumn{1}{c}{Placing} & \multicolumn{1}{c}{51.43} &
    \multicolumn{1}{c}{\textbf{73.57}} & \multicolumn{1}{c}{67.14}\\
    \multicolumn{1}{c}{Pouring} & \multicolumn{1}{c}{52.11} &
    \multicolumn{1}{c}{68.31} & \multicolumn{1}{c}{\textbf{69.01}} \\
    \midrule
    \multicolumn{1}{c}{Average} & \multicolumn{1}{c}{52.23}  &
    \multicolumn{1}{c}{\textbf{{71.30}}} & \multicolumn{1}{c}{69.16} \\
    \bottomrule 
    & & & 
    \\ 
  %\end{tabular} 
  %\begin{tabular}{@{}lc@{}lc@{}rc@{}}
    \toprule
    \multicolumn{1}{c}{{\bf AD+CD Dataset}} & \multicolumn{3}{c}{Accuracy \%}  \\
    \midrule
    \multicolumn{1}{c}{Action Class} & \multicolumn{1}{c}{BinClassifier}& \multicolumn{1}{c}{SimCLR} & \multicolumn{1}{c}{SupCLR}
     \\
    \midrule
    \multicolumn{1}{c}{Touch Nose - Eat} & \multicolumn{1}{c}{\textbf{100}} & \multicolumn{1}{c}{99.00} & \multicolumn{1}{c}{\textbf{100}}\\
    \multicolumn{1}{c}{Touch head - Shoot Ball} & \multicolumn{1}{c}{98.11} & \multicolumn{1}{c}{\textbf{100}} & \multicolumn{1}{c}{97.17}\\
    \multicolumn{1}{c}{Touch ear - Situp} & \multicolumn{1}{c}{97.30} & \multicolumn{1}{c}{\textbf{100}} & \multicolumn{1}{c}{95.95}\\
    \multicolumn{1}{c}{Tapping - Chew} & \multicolumn{1}{c}{97.91} & \multicolumn{1}{c}{\textbf{98.61}} & \multicolumn{1}{c}{96.53} \\
    \multicolumn{1}{c}{Rolly Polly - Flic flac} & \multicolumn{1}{c}{\textbf{100}} & \multicolumn{1}{c}{\textbf{100}} & \multicolumn{1}{c}{\textbf{100}}\\
    \multicolumn{1}{c}{Move the table - Push} & \multicolumn{1}{c}{99.01} &  \multicolumn{1}{c}{\textbf{100}} & \multicolumn{1}{c}{99.00}\\
    \multicolumn{1}{c}{Lock Hands - Shake Hands} & \multicolumn{1}{c}{98.82} & \multicolumn{1}{c}{98.82}  & \multicolumn{1}{c}{\textbf{100}}\\
    \multicolumn{1}{c}{Arms Up - Fall Floor} & \multicolumn{1}{c}{\textbf{98.82}} & \multicolumn{1}{c}{\textbf{98.82}} & \multicolumn{1}{c}{97.65}\\
    \midrule
    \multicolumn{1}{c}{Average} & \multicolumn{1}{c}{98.67} & \multicolumn{1}{c}{\textbf{99.34}} & \multicolumn{1}{c}{98.14}\\
    \bottomrule
  \end{tabular}
  \caption{
  ASD versus control classification results 
  (per action class and average) on the 
  %Classwise Results obtained on 
  Hand Gesture dataset (top) and the AD+CD dataset 
  (bottom). 
  %when classification task is conducted using BinClassifier, SimCLR and SupCLR. 
  }
  \label{tab:result1.1} 
  \centering
\end{table}

%\section{Result}
%\label{sec:results}
\subsection{Baseline Results}
\label{sec:baseline_results}
In this case, both the train and test data consist of samples from both the classes by creating a train-test splilt of 70:30. 
We compare the performance of binary classifier and contrastive learning methods on the 
HGD dataset in Table~\ref{tab:result1.1} (top). 
%We present comparisons between the performance of a simple binary classifier and contrastive learning methods.  
%In the case of Binary Classifier, SimCLR and SupCLR, the dataset is trained and during classification pretrained weights are used. 
We can observe that the average results 
%for HGD (complete dataset) 
improve drastically when we use contrastive learning compared to the binary classifier. 
Further, while the supervised contrastive learning method also performs better than the binary classifier, it is outperformed by the self-supervised method %results for SimCLR method improves 
by around ${2}$\%. 
%While SupCLR too performs better than Binary Classifier. 
Table~\ref{tab:result1.1} (bottom) shows the classification results on the AD+CD dataset using all the three methods. 
Based on the average accuracy, 
we can observe that self-supervised contrastive learning outperforms 
%better in classifying as compared to 
both supervised contrastive 
learning and binary classifier by 
%SimCLR performs better than both Binary Classifier and SupCLR by 
around ${1.2}\%$ and ${1.73}\%$ respectively. 
%SimCLR is better in classifying as compared to both SupCLR and BinClassifier in case of both the dataset.

%%%%%%%%%%%%%%%------TABLE 1.2------------------

%%%----------------------------------------------------------------------------
%\newline 

%--------------------------------------------------------%

\subsection{Cross-dataset Analysis} 
\label{subsec:cross_data} 

In this experiment, we perform pre-training (training of the encoder) and evaluation in a cross-dataset manner; {\it i.e.}, we perform training using the train-set of one dataset and evaluate on the test-set of another dataset. 
Table \ref{tab:result2.1} (top) shows results on 
the Hand Gesture dataset when pre-training is done on the AD+CD dataset. We observe that the accuracy obtained using self-supervised contrastive learning 
%as discussed in the previous section \ref{sec:baseline_results} 
is not affected significantly, and it performs better than the other methods. 
Similarly, Table \ref{tab:result2.1} (bottom) shows results on AD+CD dataset when pre-training is done on the Hand Gesture dataset.
%Similarly, in Table~\ref{tab:result2.1} (bottom) the encoder is pre-trained on the Hand Gesture dataset and tested on the AD+CD dataset. 
Interestingly, in case of self-supervised contrastive learning, we observe that the baseline results on the Hand Gesture dataset are slightly inferior than those obtained using cross-data training for this dataset. 
%,  while for the AD+CD dataset, the improvement is .
%%%%% Classwise ------
\begin{table}[t] %[!h]
  \centering
  \begin{tabular}{@{}lc@{}lc@{}rc@{}}
    \toprule
    \multicolumn{1}{c}{{\bf Hand Gesture Dataset}} & \multicolumn{3}{c}{Accuracy \%}  \\
    \midrule
    \multicolumn{1}{c}{Method} & \multicolumn{1}{c}{BinClassifier} & \multicolumn{1}{c}{SimCLR}  & \multicolumn{1}{c}{SupCLR}\\
    \midrule
    \multicolumn{1}{c}{Average} & \multicolumn{1}{c}{50.09} & \multicolumn{1}{c}{\textbf{73.44}}  &  \multicolumn{1}{c}{62.56} \\
    \bottomrule 
    & & & \\ 
    \toprule
    \multicolumn{1}{c}{{\bf AD+CD Dataset}} & \multicolumn{3}{c}{Accuracy \%}  \\
    \midrule
    \multicolumn{1}{c}{Method} & \multicolumn{1}{c}{BinClassifier} & \multicolumn{1}{c}{SimCLR} & \multicolumn{1}{c}{SupCLR} \\
    \midrule
    \multicolumn{1}{c}{Average} & \multicolumn{1}{c}{72.32}& \multicolumn{1}{c}{\textbf{99.74}}  & \multicolumn{1}{c}{89.80}  \\
    \bottomrule
  \end{tabular}
  \caption{Top: Results obtained on the test-set of 
  the Hand Gesture dataset when training is %conducted using BinClassifier, SimCLR and SupCLR on 
  done on the AD+CD dataset. 
Bottom: Results obtained on the test-set of 
  the AD+CD dataset when training is %conducted using BinClassifier, SimCLR and SupCLR on 
  done on the Hand Gesture dataset.   
}
  \label{tab:result2.1}
  \centering
\end{table}

\subsection{Mixed Dataset Analysis} 

In real-life applications of autism diagnosis, the data points used in training and testing may come from different distributions. 
To analyse the effectiveness of different 
methods in such situations, we perform experiments on mixed dataset setups. 
For this purpose, we synthesize three dataset setups as described below. As before, the train-test split remains 70:30 in all the experiments. 
%We named the three setups used as follows: Mixed Dataset - I, Mixed Dataset - II and Mixed Dataset - III. 

%------
\subsubsection{Setups}
\label{subsec:setups}
%We have done some experimentation on mixed datasets. For this purpose three different data setups are used as described further. The train-test split remains 70:30 throughout all the experimentation. We named the three setups used as follows: Mixed Dataset - I, Mixed Dataset - II and Mixed Dataset - III. 

\begin{enumerate} 
%\subsubsection
\item {%Mixed Dataset - I (
Setup 1:} 
In this case, the training set consists of ASD 
samples 
%\textbf{Training Set} : We use ASD samples 
from both the datasets. 
%AD+CD and HGD dataset, 
However, the control samples come only from the train split of the Hand Gesture dataset. %GD (from training split).
For the test set, the following three cases are 
considered: 
%\newline
%\textbf{Test Set} : 
\begin{itemize}
    \item Test 1: Test set of the Hand Gesture dataset with only ASD samples. %[Test 1]
    \item Test 2: Test set of the 
    AD+CD dataset.
    %[Test 2]
    \item Test 3: Combination of the test splits of both AD+CD and Hand Gesture datasets. 
    %(test split) [Test 3]
\end{itemize} 

%\subsubsection
\item {%Mixed Dataset - II (
Setup 2:} 
In this case, the training set consists of the 
%\textbf{Training Set}: We take 
ASD samples from the train split of HGD dataset, and 
ASD samples from the train+test split of the AD+CD dataset. % we take all the ASD samples (both train and test split). 
For the test set, the following case is
considered:
%\newline
%\textbf{Test Set} : 
\begin{itemize} 
\item Test 1: Test set of the Hand Gesture dataset. 
%HGD test split [Test - I] 
\end{itemize} 

%\subsubsection
\item {%Mixed Dataset - III (
Setup 3:} In this case, the training set consists of 
the merged train-splits of both AD+CD and Hand Gesture datasets from both ASD and control categories. 
%\textbf{Training Set} : We merge train split of both AD+CD and HGD dataset. 
For the test set, the following cases are 
considered: 
\begin{itemize} 
%\textbf{Test Set} : \begin{itemize}
\item Test 1: Test set of the Hand Gesture dataset. 
%HGD test set [Test1]
\item Test 2: Test set of the AD+CD dataset. 
\item Test 3: Combination of the test sets of both AD+CD and Hand Gesture datasets. %GD (test split) [Test3]
\end{itemize}

\end{enumerate} 

\subsubsection{Results} 
%In section ~\ref{subsec:setups}, all the setups are described. 
Table~\ref{tab:result2.2} shows results for the 
three setups discussed above. We can observe that 
for Setup 1, the accuracy does not improve on the Hand Gesture dataset. However, for the 
AD+CD dataset, the accuracy improves as compared to the baseline and cross-dataset results on the same test data. %dataset. 
In case of Setup 2, the results obtained are reasonable, however they are inferior to the baseline results obtained using the self-supervised learning. 
%While mixing all ASD samples from one dataset \ie both the dataset from training and test (Setup 2) performs good but could not improve the SimCLR results further.
In Setup 3, both the datasets are merged. 
For the Hand Gesture dataset, the accuracy drops significantly for the self-supervised contrastive learning method. % doesn't work. 
However, for the AD+CD dataset, the accuracy is similar to that obtained in cross-dataset evaluation. 
%for SimCLR. 
On the other hand, using supervised contrastive learning, %SupCLR for the dataset 
the accuary is comparable to that when training alone on the dataset as in Section~\ref{sec:baseline_results}, however it is better than the cross-dataset training. 
We also observe that since the third test setup (Test 3) is a combination of Test 1  and Test 2, it 
performs as expected; {\it i.e.}, approximately average of the accuracies on the two datasets. 

In general, the results in Table~\ref{tab:result1.1},~\ref{tab:result2.1} and~\ref{tab:result2.2} indicate that %self-supervised 
contrastive feature learning methods may be preferred over simple binary classifier %supervised contrastive learning 
for small data problems such as autism diagnosis, which are frequently encountered in the healthcare domain. 
%, and when the inter-class variability is small. 
%when the dataset is

\begin{table}[t] %[!h]%
  \centering
  \begin{tabular}{@{}lc@{}lc@{}rc@{}}
    \toprule
    \multicolumn{2}{c}{Dataset } & \multicolumn{3}{c}{Accuracy \%}  \\
    \midrule
    \multirow{1}{*}{Training} & \multicolumn{1}{c}{Testing} &\multicolumn{1}{c}{BinClassifier} & \multicolumn{1}{c}{SimCLR} & \multicolumn{1}{c}{SupCLR} \\
    \midrule
    \multirow{3}{*}{Setup 1}  & \multicolumn{1}{c}{Test 1} & \multicolumn{1}{c}{54.37}& \multicolumn{1}{c}{\textbf{67.02}}  & \multicolumn{1}{c}{59.53}  \\
    \multicolumn{1}{c}{} & \multicolumn{1}{c}{Test 2} & \multicolumn{1}{c}{\textbf{100}}& \multicolumn{1}{c}{\textbf{100}}  & \multicolumn{1}{c}{\textbf{100}}  \\
    \multicolumn{1}{c}{} & \multicolumn{1}{c}{Test 3} & \multicolumn{1}{c}{74.93}& \multicolumn{1}{c}{\textbf{83.25}}  & \multicolumn{1}{c}{77.77}  \\
    \midrule
    \multirow{1}{*}{Setup 2}  & \multicolumn{1}{c}{Test 1} & \multicolumn{1}{c}{56.51}& \multicolumn{1}{c}{\textbf{70.14}}  & \multicolumn{1}{c}{57.40}  \\
    \midrule
    \multirow{3}{*}{Setup 3} 
    & \multicolumn{1}{c}{Test 1} & \multicolumn{1}{c}{55.44}& \multicolumn{1}{c}{\textbf{61.50}}  & \multicolumn{1}{c}{60.96}  \\
    \multicolumn{1}{c}{} & \multicolumn{1}{c}{Test 2} & \multicolumn{1}{c}{99.07}& \multicolumn{1}{c}{\textbf{99.74}}  & \multicolumn{1}{c}{97.22}  \\
    \multicolumn{1}{c}{} & \multicolumn{1}{c}{Test 3} & \multicolumn{1}{c}{80.47}& \multicolumn{1}{c}{\textbf{82.37}}  & \multicolumn{1}{c}{81.61}  \\
    \bottomrule
  \end{tabular}
  \caption{Results for mixed dataset analysis using 
  %training for 
  various setups as discussed in 
  Section \ref{subsec:setups}.}
  \label{tab:result2.2}
  \centering
    %\vspace{-5mm}
\end{table} 

\section{Summary and Conclusion} 
\label{sec:Conclusion} 
In this paper, we have shown that a conventional deep binary classifier may not perform well on complex classification tasks such as autism diagnosis that involve small data and small inter-class variability. 
%to such an extent that it is nearly impossible to visually distinguish among classes. %Through all the experiments, whether cross-modal training or mixed data training, throughout 
On the other hand, contrastive feature learning methods can perform much better than binary classifier on such problems. 
Interestingly, while there is no difference in the encoder of these methods, the use of contrastive loss during network training brings a significant boost in the prediction accuracy. 
We further notice that unlike the conclusions drawn 
in~\cite{NEURIPS2020SupCon}, self-supervised contrastive learning performs slightly better than supervised contrastive learning on such tasks. 

%This indicates training on such a complex samples where by bare eyes its difficult to tell which samples belong to which class requires specialized training. 
%In SimCLR and SupCLR while training we generate more samples using the transformations without explicitly using more data samples it performs better with learning in the embedded space. 
%Even if we see architecture wise, there is not much difference between the three methods using same encoder. Only merging the architecture and use of contrastive loss made the difference. 

%In paper \cite{NEURIPS2020SupCon}, claims how with same architecture it perform better than SimCLR. But reverse of this through the results we can observe that between the two methods SimCLR is better in classifying than SupCLR. In our case the data we are using is different , not only that it is videos instead of the images. The dataset we are using is very small in size and also the variation between the samples of both the categories is very less. Hence, through all experiments performed it can be concluded that contrastive learning methods performs better than binary classifier. This is discussed in more details in the Appendix.

%\newpage 

%%
%% The next two lines define the bibliography style to be used, and
%% the bibliography file.
\bibliographystyle{ACM-Reference-Format}4
\nocite{*}
\bibliography{sample-sigconf}

\end{document}